\pdfoutput=1

\documentclass[11pt]{article}

\usepackage{emnlp2021}

\usepackage{times}
\usepackage{latexsym}

\usepackage[T1]{fontenc}

\usepackage[utf8]{inputenc}

\usepackage{microtype}

\usepackage{amsmath}
\usepackage{bbding}
\usepackage{amssymb}
\usepackage{arydshln}
\usepackage{booktabs}
\usepackage{multirow}

\title{RoBLEURT Submission for the WMT2021 Metrics Task}

\author{
Yu Wan$^1$\thanks{~~Work was done when Yu Wan was interning at DAMO Academy, Alibaba Group.}~~~Dayiheng Liu$^2$\thanks{~~Corresponding authors.}~~~Baosong Yang$^2$~~~Tianchi Bi$^2$~~~\textbf{Haibo Zhang}$^2$\\\textbf{Boxing Chen}$^2$~~~\textbf{Weihua Luo}$^2$~~~\textbf{Derek F. Wong}$^{1\dagger}$~~~\textbf{Lidia S. Chao}$^1$\\
$^1$NLP$^2$CT Lab, University of Macau\\
\tt{nlp2ct.ywan@gmail.com, \{derekfw,lidiasc\}@um.edu.mo}\\
$^2$DAMO Academy, Alibaba Group\\
\tt{\{liudayiheng.ldyh,yangbaosong.ybs,tianchi.btc,}\\
\tt{zhanhui.zhb,boxing.cbx,weihua.luowh\}@alibaba-inc.com}
}

\begin{document}
\maketitle
\begin{abstract}

In this paper, we present our submission to Shared Metrics Task: \textbf{RoBLEURT} (\textbf{R}obustly \textbf{O}ptimizing the training of \textbf{BLEURT}).
After investigating the recent advances of trainable metrics, we conclude several aspects of vital importance to obtain a well-performed metric model by: 1) jointly leveraging the advantages of source-included model and reference-only model, 
2) continuously pre-training the model with massive synthetic data pairs, 
 and 3) fine-tuning the model with data denoising strategy.
Experimental results show that our model reaching state-of-the-art correlations with the WMT2020 human annotations upon 8 out of 10 to-English language pairs.

\end{abstract}

\section{Introduction}

Automatically evaluating the adequacy of machine translation (MT) candidates is crucial for judging the quality of MT systems.
N-gram-based metrics, such as BLEU~\cite{papineni2002bleu}, TER~\cite{snover2006a} and chrF++~\cite{popovic2015chrf, popovic2017chrf++}, have dominated in the topic of MT metric.
Despite the success, recent studies~\cite{smith2016climbing,mathur2020tangled} also pointed out that, N-gram-based metrics often fail to robustly match paraphrases and capture distant dependencies. As MT systems become stronger in recent decades, these metrics show lower correlations with human judgements, leading the derived results unreliable.


One arising direction for metric task is using trainable model to evaluate the semantic consistency between candidates and golden references via predicting scores.
BERTScore~\cite{zhang2020bertscore}, BLEURT~\cite{sellam2020bleurt} and COMET~\cite{rei2020comet} have shown higher correlations with human judgements than N-gram-based automatic metrics.
Benefiting from the powerful pre-trained language models (LMs), e.g., BERT~\cite{devlin2019bert}, those fine-tuned metric models first derive the representation of each input, then introduce an extra linear regression module to give predicted score which describes to what degree the MT system output adequately expresses the semantic of source/reference contents.
Furthermore, related work~\cite{takahashi2020automatic,rei2020comet} reports that, metrics which additionally introduces source sentences into inputs can further boost the performance of metric model.

To push such ``model as a metric'' approach further, we present RoBLEURT -- Robustly optimizing the training of BLEURT~\cite{sellam2020bleurt}, to achieve a better consistency between model predictions and human assessments.
Specifically, for low-resource scenarios, using only hypotheses and references can give more accurate results, alleviating the sparsity of source-side language; for the high-resource language pairs, we format the model input as the combination of source, hypothesis and reference sentences, making model attending to both source input and target reference when evaluating the consistency of semantics.
Then, we collect massive pseudo data from real MT engines tagged by pseudo scores with strong baselines for supervised model pre-training.
As to the fine-tuning phase, we rescore the noisy WMT metric data of previous years with strong metric baselines, which are then utilized to fine-tune our model. 
Experimental results show that, following the setting of WMT2021 metric task, our RoBLEURT model outperforms the reported results of state-of-the-art metrics on multilingual-to-English language pairs.

\section{RoBLEURT}

\subsection{Combining Multilingual and Monolingual Language Model}
Same as previous years, translation tasks cover both low-resource and high-resource scenarios.
To give higher reliable outputs, we believe our metric model can benefit from separately pre-trained and fine-tuned over each kind of scenarios:
\begin{itemize}
    \item For low-resource multilingual-to-English language pairs, we can hardly obtain massive parallel data with high quality, nor access well-performed automatic translation systems to produce syntectic data for pre-training. We mainly consider model outputs and gold references as our model inputs. Thus we mainly consider the monolingual English language model (called RoBLEURT-\textsc{NoSrc}) in this scenario. 
    \item As to high-resource language pairs, they do not suffer from  limitations above, thus can benefit from the information of source input, model output and target reference. A multilingual version of pre-trained LM  (called RoBLEURT-\textsc{Src})  can be used for this scenario.
\end{itemize}

The main architecture of our model is ~\textsc{Transformer}~\cite{vaswani2017attention}, which has been widely used in recent researches.
As related studies point out that RoBERTa~\cite{liu2019roberta} outperforms conventional BERT~\cite{devlin2019bert}, we employ the well-trained model checkpoint from RoBERTa family.
Besides, the conventional BLEURT model is trained based on uncased-BERT, which tokenizes the input sentences with the lower case format whereas RoBERTa uses case-sensitive tokenizer, which may be helpful to distinguish more information.
Moreover, model with larger scale is generally related with better performance and higher capacity of available knowledges.

Recently, several approaches which further fine-tune RoBERTa model can give better performance over multiple natural language inference tasks. To make sure our model can also benefit from this, we finally use \texttt{RoBERTa-large-mnli}\footnote{\href{https://huggingface.co/roberta-large-mnli}{https://huggingface.co/roberta-large-mnli}} and \texttt{RoBERTa-large-xnli}\footnote{\href{https://huggingface.co/joeddav/xlm-roberta-large-xnli}{https://huggingface.co/joeddav/xlm-roberta-large-xnli}}~\cite{conneau2020unsupervised} for low-resouce and high-resource language pairs, respectively.

\subsubsection{Model Combination}
We are also interested in exploring whether we can boost the performance of combine RoBLEURT-\textsc{NoSrc} and RoBLEURT-\textsc{Src}.
Combining the outputs from models trained with different settings is widely used in MT tasks~\cite{barrault2020findings}.
In this paper, We simply use weighted combination of all available well-trained models.

\subsubsection{Input Formatting}
Our model consists of a well-trained RoBERTa model to obtain segment-level representations.
Here we also try with two solutions: the model input includes source sentence (RoBLEURT-\textsc{Src}) or not (RoBLEURT-\textsc{NoSrc}).
For the former, the model input is formatted as:
\begin{align}
    \text{<s> hyp' </s> </s> ref </s>}.
\end{align}
As the latter, due to the number of input sentences is larger than RoBERTa predefined training format, we redesigned the input format as:
\begin{align}
    \text{<s> src </s> </s> hyp' </s> </s> ref </s>}.
\end{align}

\subsubsection{Prediction Module}
To obtain a scalar value as predicted score, we directly derive the representation at the first position of input $\mathbf{X} \in \mathcal{R}^{1 \times d}$ as the representation of input tuple, where $d$ is the size of hidden states.
It is then fed to projection layer, after which we yield a scalar for describing how adequately the hypothesis express the semantics:
\begin{align}
    s = \mathbf{W} \mathbf{X}^\top + b,
\end{align}
where $\mathbf{W} \in \mathcal{R}^{1 \times d}, b \in \mathcal{R}^1$ are both trainable parameters.

During training, the learning objective is to reduce the mean squared error (MSE) between model prediction $s$ and annotated score $score$:
\begin{align}
    \mathcal{L} = (s - score) ^ 2.
\end{align}

\subsection{Continuous Pre-training with Synthetic Data}
Continuous Pre-training the model on synthetic data is proven helpful to improve the performance~\cite{sellam2020bleurt}, where BLEURT obtain the synthetic data by randomly perturbing 1.8 million segments from Wikipedia for this continuous pre-training (also called mid-training).
However, we doubt that applying datasets out of MT domain, or even use learning signals tagged from non-reliable automatic metrics (e.g., BLEU), may harm the model learning during pre-training phase.
As a consequence, we consider collecting synthetic data with real MT models over MT task datasets.
To this end, we first collect the available translation outputs by using accessible engines\footnote{We use own MT engines to obtain translation hypotheses.} to generate MT hypotheses.
Specifically, we collect high-quality cross-lingual parallel MT training data, including Czech (cs) / German (de) / Japanese (ja) / Russian (ru) / Chinese (zh) -- English (en), from the WMT News translation track of each year.
By taking the source side (cs/de/ja/ru/zh) as input for translation engines, we collect multiple triples formatted as $(src, hyp, ref)$, where $src$, $hyp$, $ref$ represent source, hypothesis, and reference respectively.

\begin{table*}
    \centering
    \small
    \begin{tabular}{lrrrrrrrrrr}
    \toprule
     \multirow{2}{*}{\textbf{Model}} & \multicolumn{5}{c}{\textbf{High-Resource}} & \multicolumn{5}{c}{\textbf{Low-Resource}} \\ 
     \cmidrule(l{2pt}r{2pt}){2-6}
     \cmidrule(l{2pt}r{2pt}){7-11}
      & \rule{0pt}{2ex}\textbf{cs} & \textbf{de} & \textbf{ja} & \textbf{ru} & \textbf{zh} & \textbf{iu} & \textbf{km} & \textbf{pl} & \textbf{ps} & \textbf{ta} \\
     \hline
     
     \multicolumn{11}{c}{\rule{0pt}{2.5ex}\textit{Baseline}} \\
     \cdashline{1-11}
     \rule{0pt}{2.5ex}\textsc{sentBLEU} & 6.8 & 41.3 & 18.8 & -0.5 & 9.3 & 18.2 & 22.6 & -2.4 & 9.6 & 16.2  \\
     TER & -4.0 & 35.5 & 4.4 & -11.7 & -1.0 & 2.1 & 12.5 & -17.2 & -3.6 & 4.6 \\
     \textsc{chrF++} & 9.0 & 43.5 & 4.4 & -11.7 & -1.0 & 24.6 & 27.5 & 3.4 & 14.5 & 18.6 \\
     BLEURT~\cite{sellam2020bleurt} & 12.6 & 45.6 & 25.8 & 9.3 & 13.7 & 25.8 & 32.7 & 5.7 & \textbf{20.7} & 23.0 \\
     COMET~\cite{rei2020comet} & 12.9 & 48.5 & 27.4 & 15.6 & 17.1 & 28.1 & 29.8 & 9.9 & 15.8 & 24.1 \\
     SOTA Results~\cite{mathur2020results} & 14.3 & 48.5 & 27.7 & 15.6 & 17.1 & 28.1 & \textbf{33.0} & 10.9 & \textbf{20.7} & 25.3 \\
     \cline{1-11}
     \multicolumn{11}{c}{\rule{0pt}{2.5ex}\textit{Our method}} \\
     \cdashline{1-11}
     \rule{0pt}{2.5ex}RoBLEURT & \textbf{15.2} & \textbf{49.3} & \textbf{29.1} & \textbf{17.3} & \textbf{17.7} & \textbf{29.0} & 31.4 & \textbf{13.2} & 20.1 & \textbf{25.4} \\
     
    \toprule
    \end{tabular}
    \caption{\textsc{daRR} Kendall correlation (\%) over WMT2020 data for each language pair (xx-en). Results of baseline systems are conducted from official report~\cite{mathur2020results}. Best viewed in bold.}
    \label{table.experiment.main}
\end{table*}

\paragraph{Adding Noise to Data}
As~\citet{sellam2020bleurt} demonstrated that, when collecting synthetic data for pre-training metric model, adding noise to data is helpful for model learning.
Due to the high quality of automatically generated MT candidates in recent decades, such noise can smoothen the distribution of semantic consistency over whole dataset, which benefits the metric model learning.
We thus follow their research, randomly select $30\%$ of collected data to be added with noise at the hypothesis side.
More specifically, we use the ``word drop'' noise -- randomly dropping words with a randomized ratio for chosen sentence -- to achieve such goal of quality reduction.
Finally, we obtain a synthetic dataset formatted as $(src, hyp', ref)$, where $hyp'$ is the noisy hypothesis.

\paragraph{Data Pseudo Labeling}
As our model tends to be a regression model -- predicting score for each inputted triplet, supervisedly guiding the model learning with given scores is essential.
To give more adequate scores for each data item, we use COMET~\cite{rei2020comet} for tagging each triplet, resulting into the data items formatting as quadruple $(src, hyp', ref, score)$.
To make sure our model should be stably trained, we rescale the scores with Z-score format following~\citet{sellam2020bleurt}.

\subsection{Fine-tuning with Data Denoising Strategy}\label{sec:Denoising}
As reported in~\newcite{sellam2020bleurt}, \newcite{ma2019results}, and \newcite{mathur2020results}, noisy data may give incorrect judgements on the reliability of one specific MT metric.
After collecting the data from previous years, we find out that the \textsc{DA} datasets from year 2018-2019 are recognized as noisy ones, however they contribute a considerable portion to the available \textsc{DA} datasets.
To give more accurate learning signals for training, we believe identifying the noisy data items is of vital importance.
Specifically, we prepare the required metrics following two methods: 
\begin{itemize}
    \item RoBLEURT checkpoints. We first train several RoBLEURT models with different portions of training data, as well as multiple experiments by setting different random seeds. Here we use both RoBLEURT and RoBLEURT-\textsc{NoSrc} settings, and derives 4 checkpoints following each setting.
    \item Available well-performed checkpoints. We collect the officially released COMET\footnote{\url{https://github.com/Unbabel/COMET}} and BLEURT checkpoints\footnote{\url{https://github.com/google-research/bleurt}}.
\end{itemize}

After collecting the predictions with all checkpoints above, we identify the noisy data items by computing the variance of rankings within whole dataset.
Finally, we rescore those noisy items with those models, tagging pseudo labels for fine-tuning.
Besides, to guarantee the scores are unbiased, we re-normalize them within the dataset of each year by Z-score following~\newcite{sellam2020bleurt}.

\section{Experiments}

\subsection{Settings of Continuous Pre-training}
\paragraph{Synthetic Data Collection}
To continue pre-training the model, we simply collect parallel data from the previous WMT conferences, taking the training data from MT track cs/de/ja/ru/zh-en language pairs to obtain high-resource pseudo data. Finally, for each language pair, we collect 2.0 million quadruples for metric model pre-training. For low-resource scenarios, we reuse the datasets above, where the only difference is removing the source sentences.

As to development set, we directly collect the direct assessment (DA) dataset from the WMT2020 Metrics task track.
We evaluate the model performance following \textsc{daRR} assessments~\cite{ma2019results,rei2020comet}, and choose the best checkpoint for fine-tuning.

\paragraph{Hyper-parameters}
During the continuous pre-training, we determine the maximum learning rate as $5\cdot10^{-6}$, training steps as 0.5M and warm-up steps as 50K.
The learning rate first linearly warms up from $0$ to maximum learning rate, then decays to $0$ till the end of training.
To avoid over-fitting, we apply the dropout ratio as 0.1.
We conduct the pre-training experiments with 8 Nvidia V100 GPUs, where each batch size for each GPU device contains 4 quadruplets.
To avoid memory issues during pre-training, we simply reduce the number of total tokens, leaving 128 and 192 for RoBLEURT-\textsc{NoSrc} and RoBLEURT-\textsc{Src}, respectively.

\subsection{Settings of Fine-tuning}
\begin{table*}
    \centering
    \small
    \begin{tabular}{lrrrrrrrrrr}
    \toprule
     \textbf{Model} & \textbf{cs} & \textbf{de} & \textbf{ja} & \textbf{ru} & \textbf{zh} & \textbf{iu} & \textbf{km} & \textbf{pl} & \textbf{ps} & \textbf{ta} \\
     \cline{1-11}
    \rule{0pt}{2.4ex}RoBLEURT-\textsc{NoSrc} & 13.5 & 46.9 & 27.4 & 10.8 & 14.8 & 28.2 & 30.6 & 8.3 & 14.7 & 25.0 \\
    RoBLEURT-\textsc{Src} & 14.1 & 47.9 & 28.7 & 11.7 & 14.9 & 27.5 & 29.9 & 6.4 & 16.0 & 24.0 \\
     \cdashline{1-11}
    \rule{0pt}{2.5ex}RoBLEURT & \textbf{15.2} & \textbf{49.3} & \textbf{29.1} & \textbf{17.3} & \textbf{17.7} & \textbf{29.0} & \textbf{31.4} & \textbf{13.2} & \textbf{20.1} & \textbf{25.4} \\
    \toprule
    \end{tabular}
    \caption{\textsc{daRR} Kendall correlation (\%) over WMT2020 data with model combination. For each setting, we present the averaged correlation with well-trained 3 models. Combining both RoBLEURT-\textsc{Src} and RoBLEURT-\textsc{NoSrc} models can give significant improvement.}
    \label{table.experiment.combination}
\end{table*}

\paragraph{Data Collection}
We fine-tune our model with the WMT2015-2019 dataset as training set, where the WMT2018-2019 subsets are processed with our data denoising strategy as discussed in \S~\ref{sec:Denoising}.
To directly confirm the effectiveness of our approach, we simply use WMT2020 dataset as dev set to compare reported results in WMT2020 metric task.

To select the model for participating the WMT2021 metric task, we divide the WMT2020 dataset into 4 folds, where the data items are firstly gathered with the identical source and reference sentence.
For each fold, we select the corresponding fold of the WMT2020 subset as the dev set, and use the combination of the WMT2015-2019 dataset and the other unused WMT2020 subsets as the training set.

\paragraph{Hyper-parameters}
During fine-tuning, we set the training steps and warm-up steps as 20K and 2K, respectively.
The other hyper-parameters are identical to those of pre-training phase.
For each fine-tuning experiment, we determine the batch size as 16, and whole training process requires one single Nvidia V100 GPU.

\paragraph{Main Results}
We first testify the effectiveness of our approach by comparing with the results from the WMT2020 Metrics Task submissions.
To be fairness, all of the model based metric baselines are trained on the WMT2015-2019 dataset.
As shown in Table~\ref{table.experiment.main}, comparing to baselines, our RoBLEURT achieves the best performance on cs/de/ja/ru/zh/iu/pl/ta-to-en settings, and achieves competitive results on km-to-en and ps-to-en.

\begin{table}
    \centering
    \small
    \begin{tabular}{lrrrrrrr}
    \toprule
     \textbf{Model} & \textbf{cs-en} & \textbf{de-en} & \textbf{ja-en} & \textbf{ru-en} & \textbf{zh-en} \\
     \cline{1-6}
     \rule{0pt}{2.4ex}base & 11.7 & 44.3 & 24.1 & 9.1 & 12.1 \\
     large & 12.4 & 46.2 & 26.2 & \textbf{12.0} & 14.1 \\
     large-xnli & \textbf{14.1} & \textbf{47.9} & \textbf{28.7} & 11.7 & \textbf{14.9} \\
    \toprule
    \end{tabular}
    \caption{\textsc{daRR} Kendall correlation (\%) over WMT2020 data with different pedestals for RoBLEURT-\textsc{Src} setting. Larger model size can give better performance for metric model, and finetuned RoBERTa-large-xnli model can push the improvement further.}
    \label{table.experiment.model_scale}
\end{table}
\section{Ablation Studies}
\subsection{Model Pedestal and Size}
We first investigate the impact of model pedestal for metric task.
As shown in Table~\ref{table.experiment.model_scale}, using RoBERTa-large instead of RoBERTa-base model as the base of RoBLEURT-\textsc{Src} model gives a better performance. Furthermore, using the fine-tuned checkpoint RoBERTa-large-xnli can further improves the performance.
This indicates our view, that powerful pre-trained LM, as well as the carefully re-optimized variants, can boost the performance of fine-tuned metric model.

\begin{table}
    \centering
    \small
    \begin{tabular}{lrrrrr}
    \toprule
     \textbf{Model} & \textbf{cs-en} & \textbf{de-en} & \textbf{ja-en} & \textbf{ru-en} & \textbf{zh-en} \\
     \cline{1-6}
      \rule{0pt}{2.4ex}w/o pretrain & 10.6 & 44.8 & 21.4 & 6.1 & 10.2 \\
      w pretrain & \textbf{14.1} & \textbf{47.9} & \textbf{28.7} & \textbf{11.7} & \textbf{14.9} \\
    \toprule
    \end{tabular}
    \caption{\textsc{daRR} Kendall correlation (\%) over WMT2020 data with data filtering. We use RoBLEURT-\textsc{Src} model to conduct the results. Simply removing the noisy portion does not help the model training. However, reintroducing them into training set after rescoring them gives a significant improvement.}
    \label{table.experiment.pretrain}
\end{table}
\subsection{Pre-training}
To identify the improvement after introducing extra pre-training steps for metric model, we conduct the results in Table~\ref{table.experiment.pretrain} for comparison.
As seen, the performance drops significantly without pre-training phase.
This caters to the previous findings~\cite{sellam2020bleurt}, where pre-training with pseudo data helps the supervised learning of metric model.

\begin{table}
    \centering
    \small
    \begin{tabular}{lrrrrr}
    \toprule
     \textbf{Model} & \textbf{cs-en} & \textbf{de-en} & \textbf{ja-en} & \textbf{ru-en} & \textbf{zh-en} \\
     \cline{1-6}
     \rule{0pt}{2.4ex}full set & 9.1 & 45.0 & 23.5 & 8.1 & 9.8 \\
      \& remove & 13.4 & 46.8 & 26.1 & \textbf{11.7} & 14.1 \\
      \& rescoring & \textbf{14.1} & \textbf{47.9} & \textbf{28.7} & \textbf{11.7} & \textbf{14.9} \\
    \toprule
    \end{tabular}
    \caption{\textsc{daRR} Kendall correlation (\%) over WMT2020 data with data filtering. We use RoBLEURT-\textsc{Src} model to conduct the results. Simply removing the noisy portion does not help the model training. However, reintroducing them into training set after rescoring them gives a significant improvement.}
    \label{table.experiment.rescore}
\end{table}
\subsection{Data Denoising Strategy}
As reported in~\cite{sellam2020bleurt}, the WMT2018-2019 DA subsets are bothered with noisy labels.
We also investigate the impact of those data, whether introducing them into model training, or even clean them via rescoring with stronger metric.
We thus arrange such ablation study during fine-tuning, and results are conducted in Table~\ref{table.experiment.rescore}.
Although the noisy portion contributes a great share of full training set (237K vs. 247K), the performance of RoBLEURT model trained without these noisy items does not diminish significantly.
After rescoring with available checkpoints, these data segments further improves model performance.

\subsection{Model Combination}
We first identify whether introducing source side information to metric model helps training.
As seen in Table~\ref{table.experiment.combination}, accepting source (row RoBLEURT-\textsc{Src}) than not (row RoBLEURT-\textsc{NoSrc}) as extra input significantly improves the correlation scores.
However, for low-resource scenarios, experimental results show that source-side information does not help much for model training.
This indicates that source information does not provide help for model training over low-resource scenarios, as the inadequacy of pre-training data may harms model training if source side is introduced. To derive better performance, one general idea is to combine several well-trained models during inference.
We also explore whether combining both RoBLEURT-\textsc{Src} and RoBLEURT-\textsc{NoSrc} models can give better performance.

As shown in Table~\ref{table.experiment.combination}, directly averaging scores from multiple models lead to a significant performance drop. On the contrary, our model, which takes models over both RoBLEURT-\textsc{NoSrc} and RoBLEURT-\textsc{Src} settings can effectively leverage the predictions, achieving significant performance gain across all language pairs.

\section{Conclusion}
In this paper, we describe our submission metric -- RoBLEURT, from the perspective of combining multilingual and monolingual language model, continuous pre-training with the massive synthetic data pairs, and fine-tuning with data denoising strategy.
Experimental results confirms the effectiveness of our pipeline, demonstrating state-of-the-art correlations with the WMT2020 human annotations upon 8 out of 10 to-English language pairs. 

\section*{Acknowledgements}
This work was supported in part by the National Key Research and Development Program of China (2018YFB1403202), the Science and Technology Development Fund, Macau SAR (Grant No. 0101/2019/A2), and the Multi-year Research Grant from the University of Macau (Grant No. MYRG2020-00054-FST).

\bibliography{emnlp2021}
\bibliographystyle{acl_natbib}

\end{document}